\documentclass[twocolumn]{article}

\pdfoutput=1

\usepackage[width=17cm,height=22cm]{geometry}
\usepackage{fancyvrb}
\usepackage{authblk}
\usepackage{hyperref}

\usepackage{amsfonts,amsmath}
\usepackage{amssymb}

\title{Distributed dictionary learning over a sensor network}
\author[1]{P. Chainais\thanks{pierre.chainais@ec-lille.fr}}
\author[2]{C. Richard\thanks{cedric.richard@unice.fr}}
\affil[1]{LAGIS UMR CNRS 8219, Ecole Centrale Lille, {\sc Sequel - INRIA} Lille, France}
\affil[2]{Laboratoire Lagrange UMR CNRS 7293, University of Nice Sophia-Antipolis, France}

\def\E{{\hbox{I\kern-.2em\hbox{E}}}}

\newcommand{\refeq}[1]{~(\ref{#1})}            
\def\argmin{\makebox{argmin}}

\newcommand{\R}{\mathbb{R}}

\newcommand{\I}{\mathbb{I}}

\newcommand{\boldpsi}{{\boldsymbol \psi}}

\newcommand{\x}{{\mathbf x}}
\newcommand{\y}{{\mathbf y}}

\renewcommand{\d}{{\mathbf d}}

\newcommand{\s}{{\mathbf s}}

\newcommand{\w}{{\mathbf w}}
\newcommand{\z}{{\mathbf z}}
\newcommand{\A}{\mathbf A}

\newcommand{\C}{\mathbf C}
\newcommand{\D}{\mathbf D}

\newcommand{\Noise}{\mathbf Z}

\renewcommand{\S}{\mathbf S}
\newcommand{\X}{\mathbf X}
\newcommand{\Y}{\mathbf Y}

\def\1{\ifmmode{\rm {I}\mkern-10.1mu
{1}\mkern0.5mu}\else{\rm {I}\kern-.56em
{1}\hskip0.5pt\ }\fi\relax}

\usepackage{graphicx} 

\graphicspath{
{../FIGURES/}
}

\begin{document}
\maketitle

\begin{abstract}
We consider the problem of distributed dictionary learning, where a set of nodes is required to collectively learn a common dictionary from noisy measurements. This approach may be useful in several contexts including sensor networks. Diffusion cooperation schemes have been proposed to solve the distributed linear regression problem. In this work we focus on a diffusion-based adaptive dictionary learning strategy: each node records observations and cooperates with its neighbors by sharing its local dictionary. The resulting algorithm corresponds to a distributed block coordinate descent (alternate optimization). Beyond dictionary learning, this strategy could be adapted to many matrix factorization problems and generalized to various settings. This article presents our approach and illustrates its efficiency on some numerical examples.
\end{abstract}

\medskip

\noindent\textbf{Keywords}: dictionary learning, sparse coding, distributed estimation, diffusion, matrix factorization, adaptive networks, block coordinate descent.

\section{Introduction}
\label{intro}

In a variety of contexts, huge amounts of high dimensional data are recorded from multiple sensors. When sensor networks are considered, it is desirable that computations be distributed over the network rather than centralized in some fusion unit. Indeed, centralizing all measurements lacks robustness - a failure of the central node is fatal - and scalability due to the needed energy and communication resources. In distributed computing, every node communicates with its neighbors only and processing is carried out by every node in the network. Another important remark is that relevant information from the data usually lives in a space of much reduced dimension compared to the physical space. The extraction of this relevant information calls for the identification of some adapted sparse representation of the data. Sparsity is an important property which favors the identification of the main components that are characteristic of the data. Each observation is then described by a sparse subset of atoms taken from a redundant dictionary. 
We study the problem of dictionary learning distributed over a sensor network in a setting where a set of nodes is required to collectively learn an adaptive sparse representation of independent observations.

Learning an adaptive representation of the data is useful for many tasks such as storing, transmitting or analyzing the data to understand its content. A basic dictionary can be obtained by using a Principal Component Analysis (PCA) also known as Karhunen-Lo\`eve decomposition in signal processing. However the number of atoms of such a decomposition is limited to the dimension of the data space. A more adapted representation is obtained by using a redundant dictionary where for instance no orthogonality property is imposed. While the number of potential characteristic sources is large, the number of effective sources which contribute to the signal observed by a sensor at a single moment is much smaller.  Many recent works have shown the interest of learning a {\em redundant dictionary} allowing for a {\em sparse representation} of the data, see \cite{tf11} for an up-to-date review. Furthermore, the problem of dictionary learning belongs to the more general family of matrix factorization problems that appear in a host of applications.

In this paper, we consider the situation where a set of connected nodes independently record data from observations of the same kind of physical system: each observation is assumed to be described by a sparse representation using a common dictionary for all sensors. For instance, a set of cameras observe the same kind of scenes or a set of microphones records the same kind of sound environment. 

The dictionary learning and the matrix factorization problems are connected to the linear regression problem. Let us consider that a set of observations of the system is described by a data matrix $\S$ where each column corresponds to one observation.  Assume that  $\S = \D\X$. If either the coefficients $\X$ (resp. the dictionary $\D$)  are known, the estimation of the dictionary (resp. the coefficients) knowing $\S$ is a linear regression problem. It appears that several recent works have proposed efficient solutions to the problem of least mean square (LMS) distributed linear regression, see \cite{cs10} and references therein. The main idea is to use a so-called {\em diffusion strategy}: each node $n$ carries out its own estimation $\D_n$ of the same underlying linear regression vector $\D$ but can communicate with its neighbors as well. The information provided to some node by its neighbors is taken into account according to some weights interpreted as diffusion coefficients. Under some mild conditions, the performance of such an approach in terms of mean squared error is similar to that of a centralized approach \cite{zs12}. Denoting by $\D_c$ the centralized estimate which uses all the observations at once, it can be shown that the error $\E\| \D_n-\D\|_2$ of the distributed estimate is of the same order as $\E\| \D_c-\D\|_2$: diffusion networks match the performance of the centralized solution. 

Our work gives strong indication that the classical dictionary learning technique based on block coordinate descent on the dictionary $\D$ and the coefficients $\X$ can be adapted to the distributed framework by adapting the diffusion strategy mentionned above. Our numerical experiments also strongly support this idea. The theoretical analysis is the subject of ongoing work. Note that solving this type of matrix factorization problems is really at stake since it corresponds to many inverse problems: denoising, adaptive compression, recommendation systems... A distributed approach is highly desirable both for use in sensor network and for parallelization of numerically expensive learning algorithms.


The paper is organized as follows. Section~\ref{problem} formulates the problem we are considering. Section \ref{dicolearning} recalls about dictionary learning techniques based on block coordinate descent approaches.  Section~\ref{distrib_dicolearning} presents the diffusion strategy for distributed dictionary learning. Section \ref{numeric_exp} shows some numerical experiments and results. Section \ref{conclusion} points to main claims and prospects. 

\section{Problem formulation}
\label{problem}
Consider $N$ nodes over some region. In the following, boldfaced letters denote column vectors, and capital letters denote matrices. The node $n$ takes $q_n$ measurements $\y_n(i)$, $1\leq i\leq q_n$ from some physical system. All the observations are assumed to originate from independent realizations $\s_n(i)$ of the same underlying stochastic source process $\s$. Each measurement is a noisy measurement 
\begin{equation}
\label{meas_model}
	\y_n(i) = \s_n(i) + \z_n(i)
\end{equation}
where $\z$ denotes the usual i.i.d. Gaussian noise with covariance matrix $\Sigma_n = \sigma_n^2\I$. Our purpose is to learn a redundant dictionary $\D$ which carries the characteristic properties of the data. This dictionary must yield a sparse representation of $\s$ so that:
\begin{equation}
\label{sparse_rep}
	\forall n, \quad \y_n(i)  =  \underbrace{\D\x_n(i)}_{\s_n(i)} + \z_n(i)
\end{equation}
where $\x_n(i)$ features the coefficients $x_{nk}(i)$ associated to the contribution of atom $\d_k$, the $k$-th column in the dictionary matrix $\D$, to $\s_n(i)$. The sparsity of $\x_n(i)$ means that only few components of $\x_n(i)$ are non zero. 

We are considering the situation where a unique dictionary $\D$ generates the observations at all nodes. On the contrary, observations will not be shared between nodes (this would be one potential generalization). Our purpose is to learn (estimate) this dictionary in a distributed manner thanks to in-network computing only, see section~\ref{distrib_dicolearning}. As a consequence, each node will locally estimate a local dictionary $\D_n$ thanks to i) its observations $\y_n$ and ii) communication with its neighbors. The neighborhood of node $n$ will be denoted by ${\cal N}_n$, including node $n$ itself. The number of nodes connected to node $n$ is the degree $\nu_n$. 

\section{Dictionary learning strategies}
\label{dicolearning}

\subsection{Problem formulation}
Various approaches to dictionary learning have been proposed \cite{tf11}. 
Usually, in the centralized setting, the $q$ observations are denoted by $\y(i)\in \R^p$ and grouped in a matrix $\Y=[\y(1),...,\y(q)]$. As a consequence, $\Y\in\R^{p\times q}$. The dictionary (associated to some linear transform) is denoted by $\D\in \R^{p\times K}$: each column is one atom $\d_k$ of the dictionary. 
We gather the coefficients associated to  observations in a single matrix $\X=[\x(1),...,\x(q)]$. We will consider learning methods  based on block coordinate descent or alternate optimization on $\D$ and $\X$ with a sparsity constraint on $\X$ \cite{t01,tf11}.

The data is represented as the sum of a linear combination of atoms and a noise term $\Noise\in \R^{p\times q}$:
\begin{equation}
\label{datadicocoeff}
	\Y = \D \X + \Noise
\end{equation}
Since dictionary learning is a {\em matrix factorization} problem, it is an ill-posed problem. The dictionary is potentially redundant and not necessarily orthogonal so that $K\gg p$. Some modeling is necessary to constrain the set of possible solutions. Various conditions can be considered (non-negative matrix factorization, orthogonal decomposition, ...). In general, a dictionary is considered as {\em adapted} to the data if each observation $\y(i)$ can be described by a little number of coefficients $\x(i)$. One usually searches for a {\em sparse representation}  and imposes the sparsity of $\X$ \cite{tf11}.

\subsection{Learning a redundant dictionary for sparse representation}
The properties of redundancy of the dictionary and sparsity of the coefficients are complementary. The extreme case would be the one where the dictionary contains each one of the true data underlying the noisy observation so that only one non zero coefficient in $\x(i)$ would be sufficient to describe the observation $\y(i)$. Then we would have $K=q$ and $\X$ would be maximally sparse (only 1 non zero coefficient per observation). Of course this dictionary would not be very interesting since its generalization power would be very limited. A good dictionary must offer a compromise between its fidelity to the learning data set and its ability to generalize. The choice of the size of the dictionary is often made a priori so that $K>p$ to ensure some redundancy and $K<q$ to ensure it can capture some general information shared by the data. For instance, when working on image patches of size $8\times 8$ (the data lives in dimension $p=64$), it is typically proposed to learn dictionaries of size $256$ or $512$ \cite{aeb06,tf11}.

In the classical setting, the noise is usually assumed to be i.i.d. Gaussian noise so that the reconstruction error is measured by the L2-norm. Sparsity of the coefficient matrix is imposed through a L0 relaxed to L1-penalization in the mixed optimization problem:
\begin{equation}
\label{L1penaltypb}
	(\D,\X) = \argmin_{(\D,\X)} \quad\frac{1}{2} || \Y-\D\X||_2^2 + \lambda ||\X||_1
\end{equation}
Under some mild conditions, this problem is known to provide a solution to L0-penalized problem (ideally we would prefer to directly solve the L0-penalized problem) \cite{smf10}.

\subsection{Block coordinate descent}
\label{blockdescent}

One way to solve problem\refeq{L1penaltypb} is to use block coordinate descent \cite{t01}, that is alternate optimization on $\X$ and $\D$. There are several possibilities to do this, see e.g. \cite{aeb06}. For instance, after some initialization, one may use gradient descents on $\X$ and $\D$ \cite{of96}. Such approaches are attractive since we know that linear regression by gradient descent can be translated in the distributed framework \cite{cs10}. 

One possible choice is the Basis Pursuit algorithm. At each step, the forward-backward splitting (Basis Pursuit Denoising with Iterated Soft Thresholding, see \cite{smf10}, p.161) iteratively estimates $\X$ by iterating the following steps over $s$ and $t$:
	\begin{enumerate}
		\item $\quad \X^{(s, t+1/2)} = \X^{(s,t)} + \lambda\mu  \D^{(s,t)T} \left[ \Y-\D^{(s)}\X^{(s,t)}\right] $\\ (gradient descent step, $\forall n$)
		\item $\X^{(s,t+1)} = \mbox{SoftThreshold}_{\lambda\mu}(\X^{(s,t+1/2)})$  \qquad\qquad(soft thresholding step)
	\end{enumerate}
	Then we update $\X^{(s+1)} = \X^{(s,T)}$ after $T$ (typically 30 or 40) iterated soft thresholding.
Note that one must have $\mu \in \left(0,\frac{2}{||\D||_F^2}\right)$ where $||\cdot||_F$ denotes the Frobenius norm. 
Then the dictionary $\D^{(s+1)}$ can be updated knowing $\X^{(s+1)}$, using a simple gradient descent:
\begin{equation}
\label{dico_update_GD}
	\tilde{\D}^{(s+1)} = \D^{(s)} + \eta\left[\Y - \D^{(s)}\X^{(s+1)} \right]\X^{(s+1)T}
\end{equation}
which tends to minimize $||\Y-\D\X||^2_F$ with respect to $\D$ for $0<\eta<2/\lambda_{\mbox{max}}(\X^T\X)$ ($\lambda_{\mbox{max}}$ stands for the largest eigen value.
The dictionary is then normalized:
\begin{equation}
\label{normald}
		\forall \leq k\leq K, \d_k = \frac{1}{\|\tilde{d}_k\|_2}\; \tilde{\d}_k. 
\end{equation}
One may also use Moore-Penrose pseudo-inverse following the MOD \cite{eah99}:
\begin{eqnarray}
	\tilde{\D}^{(s+1)} & = & \argmin_{\D} \quad\frac{1}{2} || \Y-\D\X^{(s+1)}||_2^2 \nonumber\\
	& = & \Y \X^{(s+1)T}\cdot \left(\X^{(s+1)}\X^{(s+1)T}\right)^{-1} \label{dico_pseudoinv}
\end{eqnarray}
again followed by a normalization step. Other more sophisticated methods have also been proposed like FOCUSS \cite{mk01}, K-SVD \cite{aeb06} or the majorization method \cite{ybd09}.	We do not discuss all these methods here for sake of briefness. In the following, it appears that methods rooted in the simple gradient descent update is the easiest to adapt to the distributed diffusion strategy. The comparison of performances of various methods is under study.

\section{Distributed dictionary learning}
\label{distrib_dicolearning}

\subsection{Diffusion strategies for distributed estimation}
\label{distrib_estim}

This section presents one particular effective diffusion strategy to solve LMS distributed estimation problems, see \cite{cs10,zs12} for a detailed presentation. Here we focus on the Adapt-Then-Combine (ATC) strategy.

The Adapt-Then-Combine (ATC) strategy aims at solving the problem of a {\em scalar} least mean squares linear regression over a sensor network. Observations $y_n(i)$ are assumed to arrive sequentially at consecutive instants $i$. In the usual setting \cite{cs10}, each sensor records both a noisy {\em scalar} measurement $y_n(i)\in\R$ and a set of coefficients $\x_n(i)$ under the assumption 
\begin{equation}
\label{basicreg}
	y_n(i) = \w_o^{T}\x_{n,i} + z_n(i).
\end{equation}
The objective is to collectively estimate $\w_o$. The purpose of ATC is that the sensors $\{ n : 1... N\}$ yield estimates $\w_n$ of the common underlying regression vector $\w_o$ from observations $\{y_n(i) ; \x_n(i)\}$ at time $i$. The cost function under the assumption of Gaussian noise is:
\begin{equation}
\label{basiccost}
	J(\x,\w) =\sum_{n=1}^N \underbrace{\E |y_n(i)-\w^T\x_{n,i}|^2 }_{J_{loc}(\x_{n,i},\w)}
\end{equation}
Let $\A,\; \C\in (\R^+)^{N\times N}$ two matrices such that:
\begin{equation}
\label{def_coeffAC}
\left\{
\begin{array}{l}
	c_{\ell,n} = a_{\ell,n} = 0 \mbox{ if } \ell\notin{\cal N}_n,\\[2mm]
	{\mathbf 1}^T\C =  {\mathbf 1}^T, \C {\mathbf 1}= {\mathbf 1},  {\mathbf 1}^T\A =  {\mathbf 1}^T	
\end{array}
\right.
\end{equation}
where ${\mathbf 1}$ is column vector of ones. The ATC algorithm consists of 2 steps:
\begin{eqnarray} 
	\boldpsi_{n,i} & =  & \w_{n,i-1} +  \quad\qquad (Adapt) \label{adapt}\\
	& &\mu_n^w \sum_{\ell\in{\cal N}_n} c_{\ell,n}^w \underbrace{\x_{\ell,i-1}[y_\ell(i) -  \w_{n,i-1}^T\x_{\ell,i-1}]}_{\nabla_w J_{loc} (\x_{\ell,i-1},\w_{n,i-1})}\nonumber\\
	\w_{n,i} & = & \sum_{\ell\in{\cal N}_n} a_{\ell,n}^w \boldpsi_{\ell,i}   \qquad (Combine) \label{combine}
\end{eqnarray}
The ATC algorithm can be seen as a distributed gradient descent where each sensor tries to estimate $\w^o$ as $\w_{n,i}$ by exploiting its own measurement $y_n(i)$ as well as information shared with its neighbors.  Eq.\refeq{adapt} is the {\em Adapt} or {\em incremental} step, eq.\refeq{combine} is the {\em Combine} or {\em diffusion} step which averages estimates from neighbors of node $n$. As a consequence, a local (possibly averaged if $\C\neq \I$) gradient with respect to $\w$ is computed at each node. An intermediate updated version of the local estimate of $\w_o$ denoted by $\psi_{n,i}$ is then obtained. The final estimate at each node is a local average of neighboring intermediate estimates. 

In the sequel, we will focus on the case where observations are not shared between nodes so that matrix $\C = (c_{\ell,n})$ is simply identity $\C=\I$. Various choices can be considered for $\A$. In the numerical experiments below we typically work with either some a priori fixed matrix $\A$ or with the relative degree variance:
\begin{equation}
\label{relatdegvar}
	a_{\ell,n} = \frac{\nu_\ell \sigma_\ell^2}{\sum_{m\in{\cal N}_n}\nu_m \sigma_m^2}
\end{equation} 

The performance analysis of this ATC diffusion strategies and some other variants can be found in \cite{zs12}. The mean-square error of the ATC estimate of $\w_o$ is similar to that of the centralized version (which would see all the observations at once). As a conclusion, this diffusion strategy is very powerful to deal with a distributed solution to a linear regression problem. Let us emphasize that in this setting each observation is made of a couple $(y_n,\x_n)$ where $y_n$ is a scalar. In the dictionary learning problem, only the {\em vector} $\y_n$ will be observed and both the dictionary (therefore $\D$ in place of $\w_o$) and the coefficient $\x_n$ are to be jointly estimated: this is a factorization problem.

\subsection{Distributed alternate optimization for dictionary learning}
\label{distrib_altoptim}

The ATC diffusion strategy for distributed estimation described above originates the following approach to distributed block-coordinate descent (alternate optimization) for dictionary learning. We will mainly keep the concept of {\em diffusion} to ensure communication between nodes: every node will share its dictionary estimate with its neighbors in ${\cal N}_n$.
Let us remark some differences in our setting compared to setting of section~\ref{distrib_estim}. Observations will be the {\em vectors} (not only scalar) $\y_n(i), i=1...q_n$ at node $n$. Observations are taken simultaneously at each node, not sequentially, so that a whole data matrix $\Y_n$ is assumed to be available at node $n$. Here index $i$ stands for iterations. The case where data arrive sequentially at each node can also be dealt with at the price of a natural adaptation of the present approach. Note that the $\x_{n,i}$ are not known anymore: each node must estimate both its local dictionary $\D_n$ and the coefficients $\X_n$ which describe observations $\Y_n = \D_n\X_n + \Noise_n$. At each iteration $i$, only the local dictionary estimates $\D_{n,i}$ are assumed to be shared between neighbors, not observations, so that $\C=\I$ in eq.\refeq{def_coeffAC}.
\begin{table}[htcb]
\begin{tabular}{l}
	\hline
	Algorithm 1: ATC for sparse dictionary learning\\ 
	\hline
	Initialize $\D_{n,0}, \:\forall n$ (see in the text for various options).\\
	Given a matrix $\A$	satisfying~\refeq{def_coeffAC}, $i=0$, \\[2mm]
	
	 {\bf Repeat until convergence of $(\D_{n,i},\X_{n,i})_{n=1:N}$}\\[2mm]

	 {\bf For each node $n$ repeat:}\\[2mm]
	 	
	1) Optimization w.r.t. $\X_{n,i}$ (sparse rep.):\\
		Given the dictionary, the coefficients are \\ 
		iteratively updated  through \\
		\hspace*{2mm}{\bf For $t=1:M$} (typically $M=30$)\\
		\hspace*{4mm} i) $\X_{n,i}^{(t+1/2)} = \X_{n,i}^{(t)} +$
		$\lambda_n\mu_n^X  \D_{n,i}^T(\Y_n-\D_{n,i}\X_{n,i}) $\\ 
		\hfill (gradient descent step)\\
		\hspace*{4mm} ii) $\X_{n,i}^{(t+1)} = \mbox{SoftThreshold}_{\lambda_n\mu_n^X}(\X_{n,i}^{(t+1/2)})$\\
		\hspace*{2mm}{\bf EndFor ($t$)}\\[2mm]

	2) Optimization w.r.t. $\D_{n,i}$ (dictionary):\\
		$
		\left\{
		\begin{array}{rcl}
			\boldpsi_{n,i+1} & = & \D_{n,i} + \mu_n^D (\Y_n-\D_{n,i}\X_{n,i})\X_{n,i}^T\\[2mm]
			\D_{n,i+1} & = & \sum_{\ell\in{\cal N}_k} a_{\ell,n}^D \boldpsi_{\ell,i} \mbox{ (diffusion)}
		\end{array}
		\right.
		$\\[6mm]
				
		{\bf EndFor ($n$)}\\[2mm]
		$i\leftarrow i+1$\\
	 {\bf EndRepeat}\\

	\hline
\end{tabular}
\end{table}
The algorithm 
goes as follows. First the local dictionaries $\D_{n,0}$ are initialized to a random set of $K$ observations (columns) from $\Y_n$ at node $n$.  Then we iteratively solve the sparse representation problem~\refeq{L1penaltypb} at each node, for instance using the forward-backward splitting method over a large number $M$ of iterations, see section~\ref{blockdescent}. The penalty parameter $\lambda_n$ may be adjusted for each node according to the local noise level $\sigma_n^2$. Note that positive learning rates $\mu_n^{X}$ (resp. $\mu_n^{D}$) must obey the condition $\mu_n^{X}<\frac{2}{||\D_{n,i}||_F}$ (resp. $\mu_n^{D} <\frac{2}{||\X_{n,i}||_F}$).

In summary, each node updates its dictionary as a function of its local observations $\Y_n$ (Adapt step) and its neighbors' dictionaries (Combine step). Sparse representations are computed locally. Based on known results for the ATC strategy in its usual setting, we expect the present Algorithm 1 above 
converges to an accurate estimate of the common underlying dictionary $\D$. Next section supports this intuition thanks to numerical experiments on images.

\section{Numerical experiments \& results}
\label{numeric_exp}

We present some numerical experiments to illustrate the relevance and efficiency of our approach. In the spirit of the seminal work by Olshausen \& Field \cite{of96} our algorithm was tested on datasets containing controlled forms of sparse structure. We consider a set of $r\times r$ image patches composed of sparse pixels. Each pixel was activated independently according to an exponential distribution, $P(x) \propto e^{-|x|}$. 

\begin{figure}
 {\setlength{\tabcolsep}{0pt}
 \begin{tabular}{cc}
	\includegraphics[width=40mm]{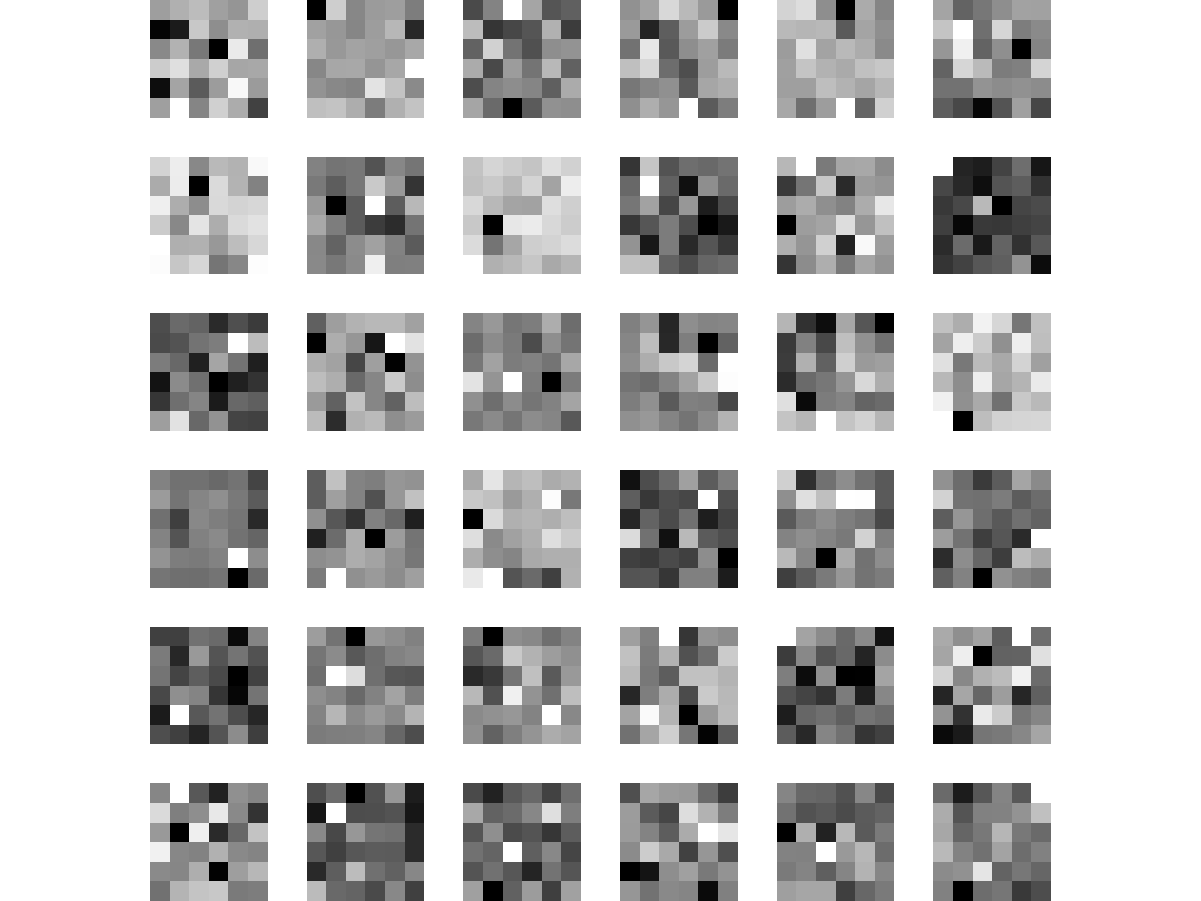} 
	&
	\includegraphics[width=40mm]{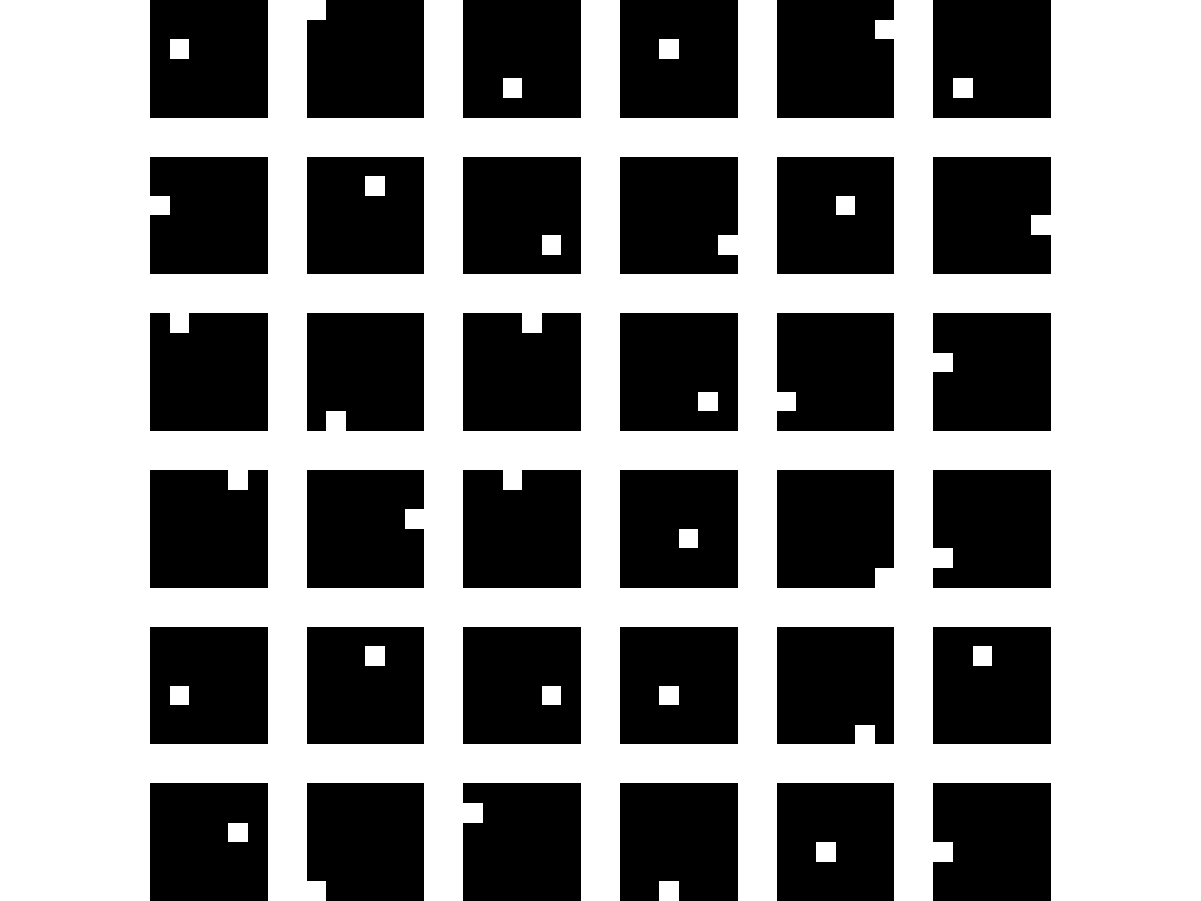} \\
	(a) & (b)\\
	\includegraphics[width=40mm]{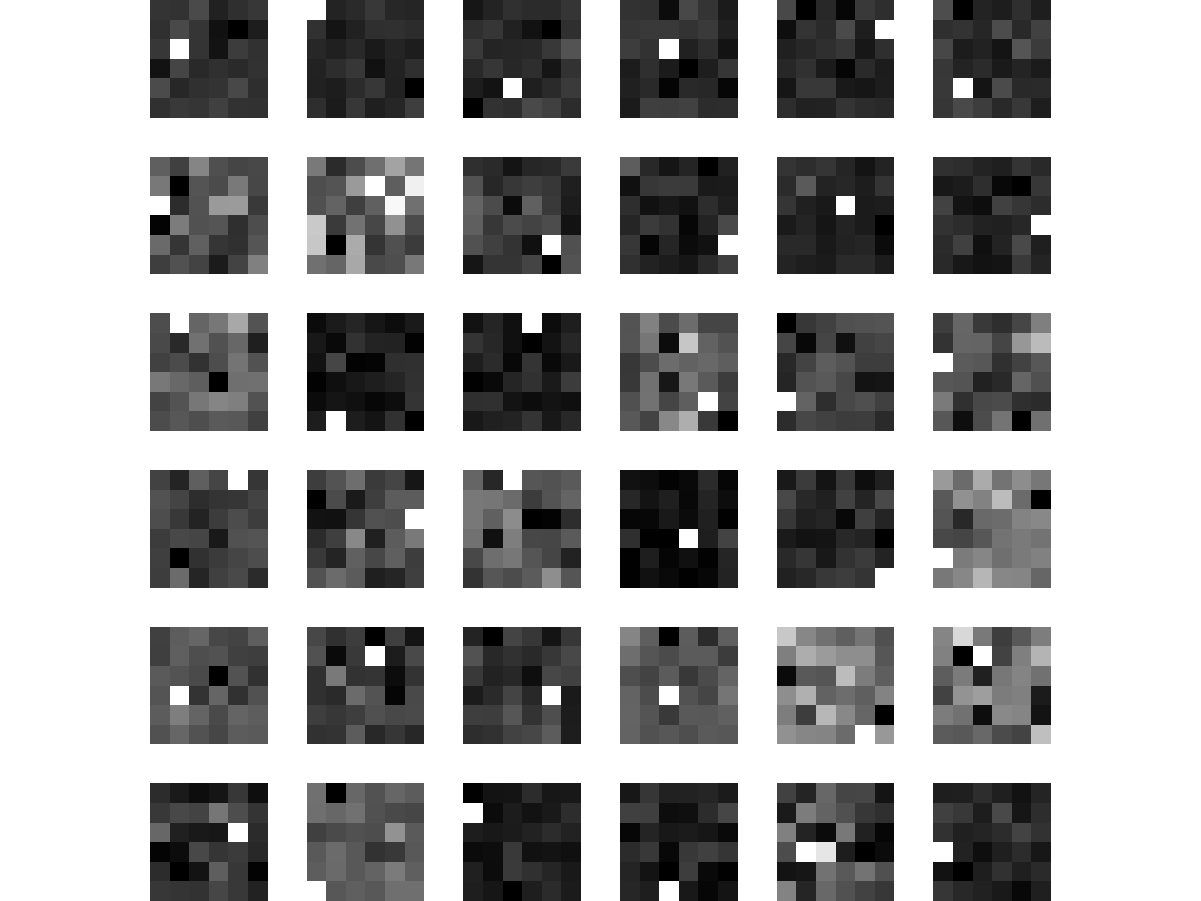} 
	&
	\includegraphics[width=40mm]{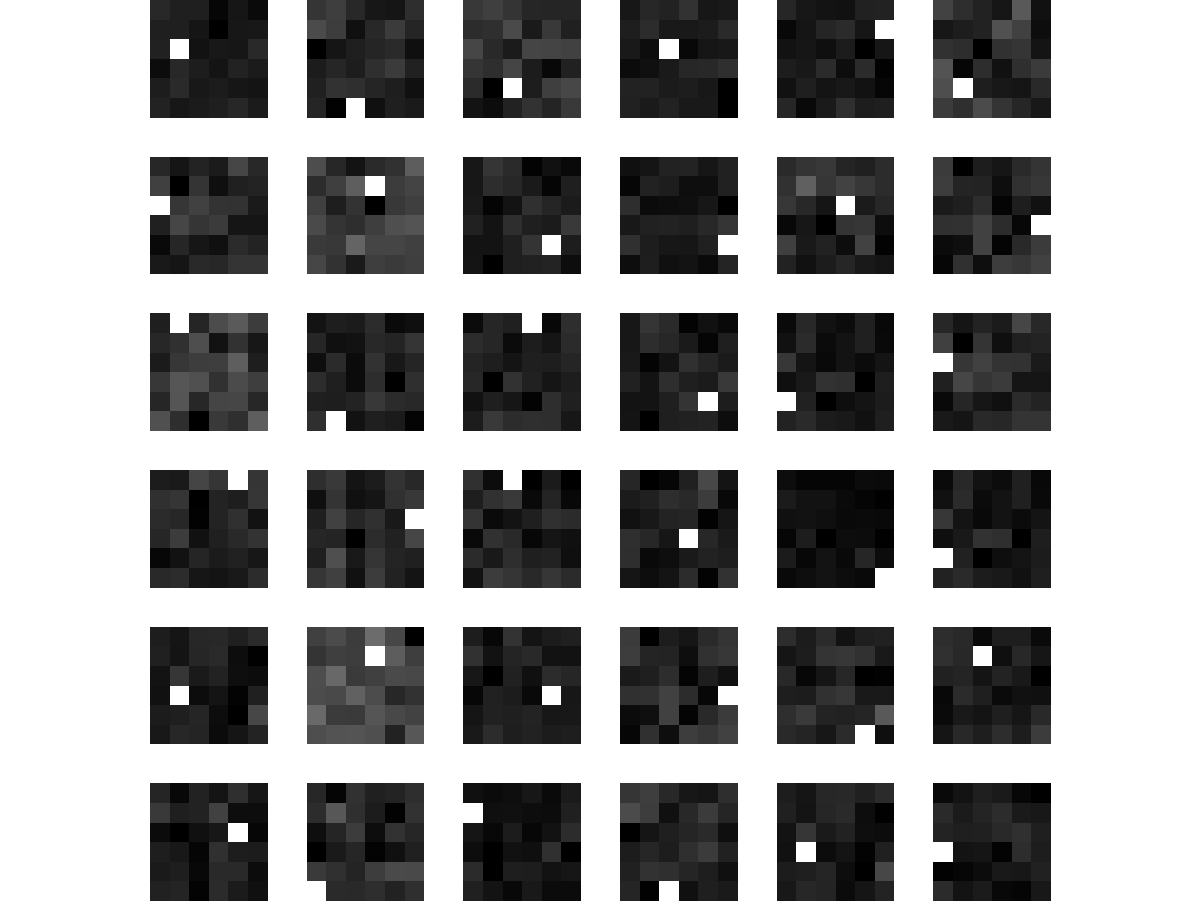}\\
	(c) & (d)\\
	\includegraphics[width=40mm]{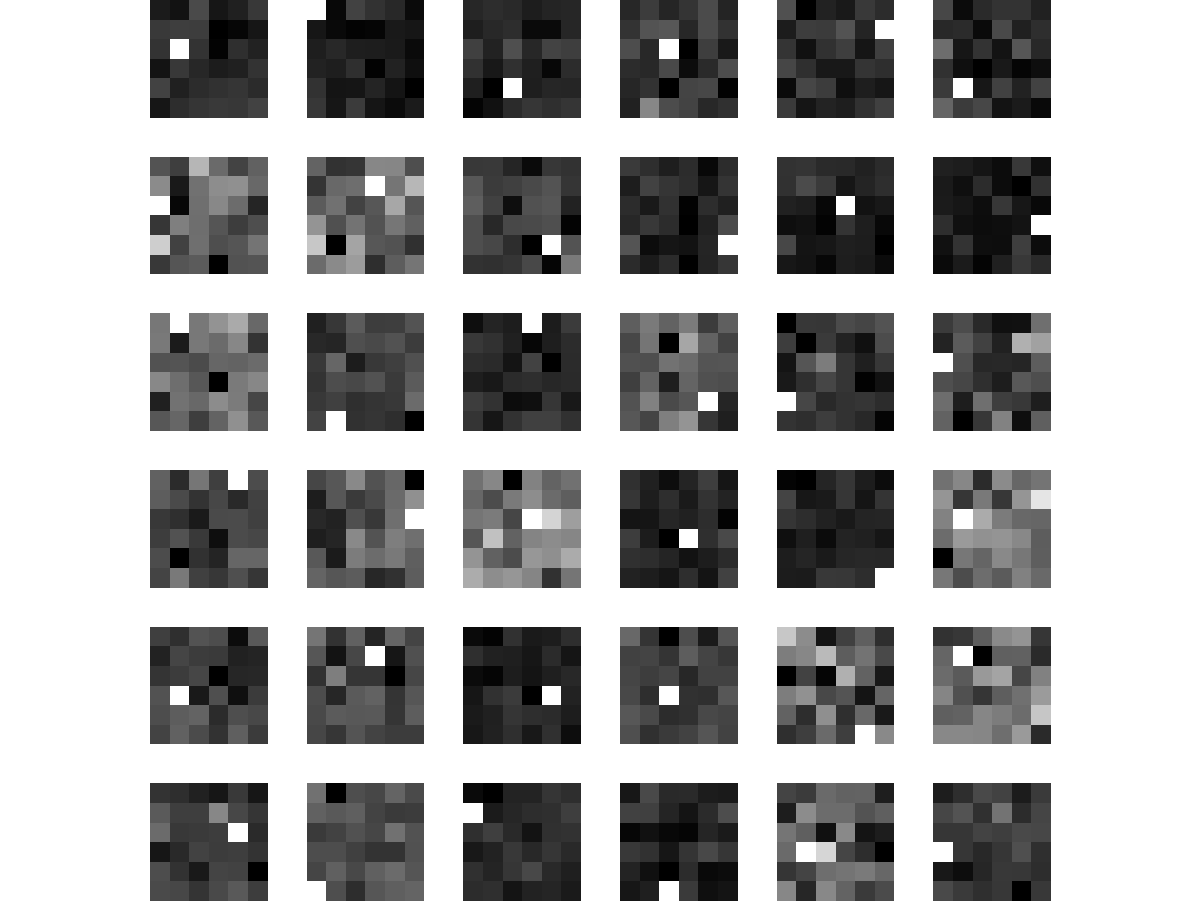} 
	&
	\includegraphics[width=40mm]{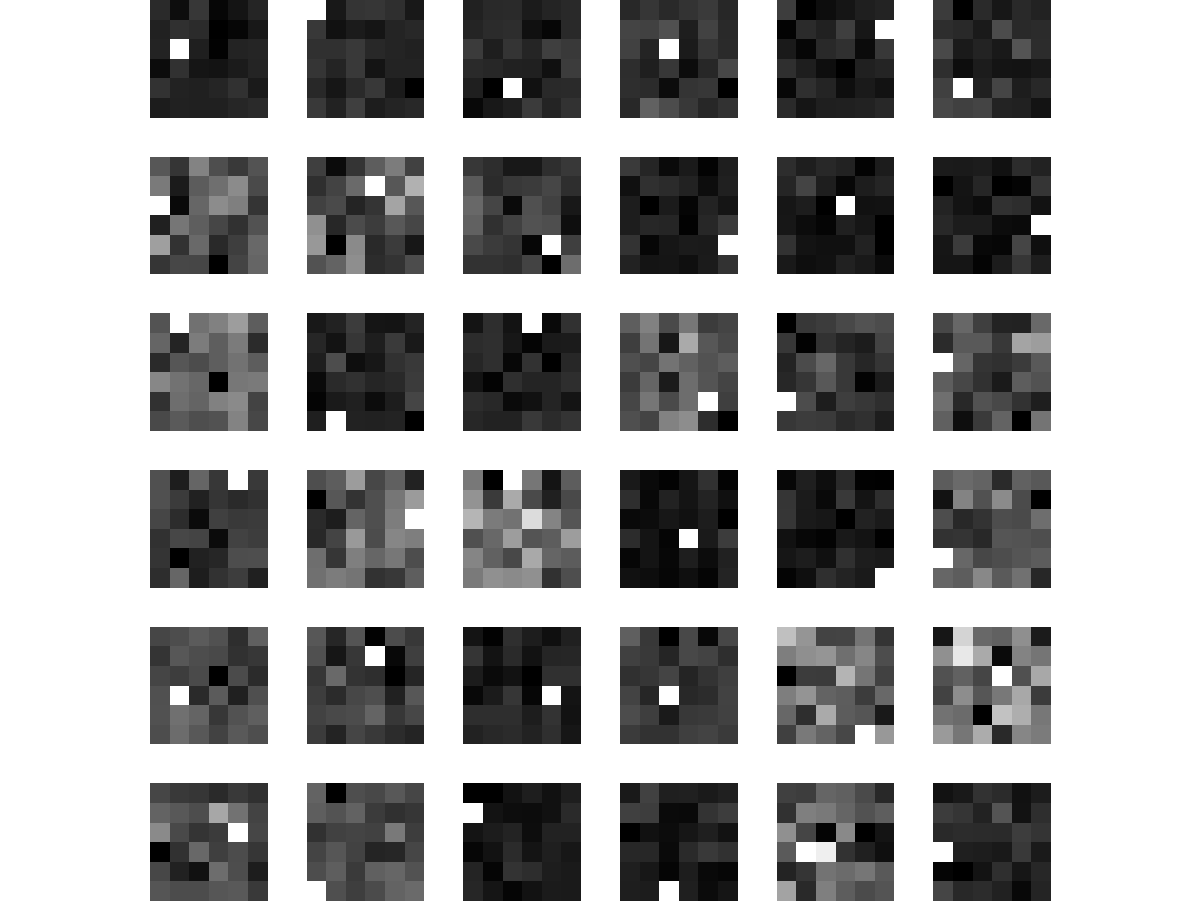} \\
	(e) & (f)
\end{tabular}}
\caption{\label{figres1} (a) Examples of patches, (b) true dictionary used for synthesis, (c) \& (e) local dictionaries for 2 different nodes, (d) dictionary learnt from the usual centralized learning, (f) dictionary averaged over nodes' estimates. Atoms have been reordered to make comparisons easier.}
\end{figure}

We consider the simple situation of a set of 4 nodes in a symmetrically connected network. Thus we used a symmetric matrix $\A$ such that:
\begin{equation}
\label{A1}
\A = \left[
\begin{array}{cccc}
	0.6 & 0.2 & 0 & 0.2\\
	0.2 & 0.6 & 0.2 & 0\\
	0 & 0.2 & 0.6 & 0.2\\
	0.2 &  0   & 0.2 & 0.6
\end{array}
\right]
\end{equation}
Note that nodes are not even directly connected one to all the others.

Fig.~\ref{figres1} shows that all the nodes have consistently learnt the same dictionary. Let us emphasize that these dictionaries are consistent, in the sense that no local reordering was necessary at any step. All the nodal dictionaries $\D_n$ are close to the same common dictionary $\D$. It appears that the mean-square error over all estimates is similar to that obtained from the centralized dictionary learning procedure of section~\ref{dicolearning}. Therefore even though each nodeslocally solves a matrix factorization problem from a particular disjoint subset of observations, the same common dictionary is (approximately) identified. This is made possible by the diffusion principle which relies on a simple communication between neighbors only.

\section{Conclusion \& Prospects}
\label{conclusion}

As a conclusion, we have presented an original algorithm which solves the problem of distributed dictionary learning over a sensor network. This is made possible thanks to a diffusion strategy which permits some local communication between neighbors. Connected nodes can exchange their local dictionaries which are estimated from disjoint subsets of data. This algorithm is the adaptation of usual dictionary learning techniques for sparse representation to the context of in-network computing. Some numerical experiments illustrate the relevance of our approach. The theoretical study of the algorithm is the subject of ongoing work. 
Several improvements and generalizations can also be considered. Many methods are available for sparse coding. This choice is crucial to get better dictionary estimates. We will study which method is most adapted to this distributed setting. The optimization of communication coefficients may be of some help as well. 

We believe that this approach to the general problem of distributed matrix factorization opens the way towards many prospects and applications. Moreover, as far as computational complexity is concerned, distributed parallel implementations are a potentially interesting alternative to online learning techniques \cite{mbps10}. 

\bibliography{cap13_biblio}

\end{document}